\documentclass[conference]{IEEEtran}
\IEEEoverridecommandlockouts
\usepackage{cite}
\usepackage{amsmath,amssymb,amsfonts}
\usepackage{graphicx}

\usepackage{algorithm2e}
\usepackage{textcomp}
\usepackage{xcolor}
\usepackage{graphicx}
\usepackage{comment}
\usepackage[bookmarks=false]{hyperref}
\usepackage{color,soul}
\def\BibTeX{{\rm B\kern-.05em{\sc i\kern-.025em b}\kern-.08em
    T\kern-.1667em\lower.7ex\hbox{E}\kern-.125emX}}
\begin{document}
\IEEEaftertitletext{\vspace{-1\baselineskip}}
\title{Fusing Frame and Event Vision for High-speed Optical Flow for Edge Application\\}

\vspace{5em}
\IEEEaftertitletext{\vspace{-1\baselineskip}}
                
        
\author{\IEEEauthorblockN{Ashwin Sanjay Lele, Arijit Raychowdhury}
\IEEEauthorblockA{\textit{School of Electrical and Computer Engineering, Georgia Institute of Technology, Atlanta, GA, USA} \\
alele9@gatech.edu, arijit.raychowdhury@ece.gatech.edu}
}

\vspace{-3mm}

\maketitle

\begin{abstract}
Optical flow computation with frame-based cameras provides high accuracy but the speed is limited either by the model size of the algorithm or by the frame rate of the camera. This makes it inadequate for high-speed applications. Event cameras provide continuous asynchronous event streams overcoming the frame-rate limitation. However, the algorithms for processing the data either borrow frame like setup limiting the speed or suffer from lower accuracy. We fuse the complementary accuracy and speed advantages of the frame and event-based pipelines to provide high-speed optical flow while maintaining a low error rate. Our bio-mimetic network is validated with the MVSEC dataset showing 19$\%$ error degradation at 4$\times$ speed up. We then demonstrate the system with a high-speed drone flight scenario where a high-speed event camera computes the flow even before the optical camera sees the drone making it suited for applications like tracking and segmentation. This work shows the fundamental trade-offs in frame-based processing may be overcome by fusing data from other modalities.

\end{abstract}


\vspace{-0.5cm}
\section{Introduction}

Computation of optical flow (OF) finds applications in many computer vision and robotics tasks ranging from pose estimation \cite{pose_estimation}, video stabilization \cite{video_stabilization}, visual odometry \cite{flow_odometry}, collision avoidance \cite{collision_avoidance} to feature tracking \cite{feature_tracking} etc. Consistent previous exploration into both model-based algorithms \cite{model_based_survey} and convolutional neural networks (CNNs) \cite{cnn_survey} have achieved unsurpassed accuracy levels for this task. Highly accurate CNNs require significant inference latency \cite{flowfields} while smaller models \cite{liteflownet} trade off accuracy for speed. Model-based optimization techniques \cite{farneback_speed} require significant computation time due to a large number of memory accesses and pipelined computations. Even faster methods eventually get limited by the frame rate of the regular optical camera so reducing the camera resolution to accelerate the computation is not enough. Therefore the optical cameras along with conventional computer vision techniques remain inadequate for high-speed edge applications due to discrete data processing modality. \footnote{To appear in the proceedings of IEEE International Symposium on Circuits and Systems (ISCAS) 2022}

\begin{figure}[h!]
\centerline{\includegraphics[width=0.5\textwidth,height=0.61\textwidth]{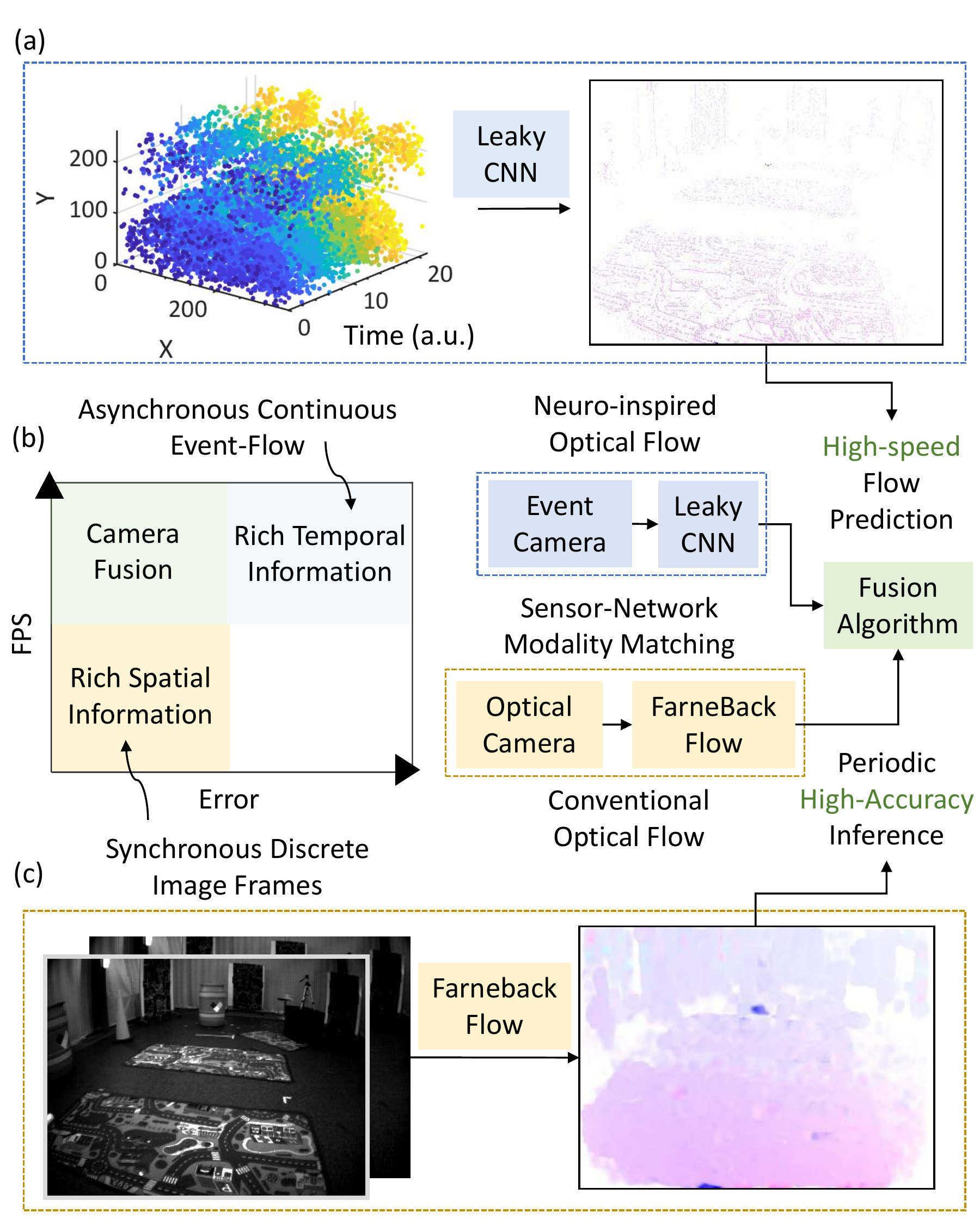}}
\centering
\vspace{-0.2cm}
\caption{ (a) High speed OF computation using proposed leaky CNN filter for event camera (c) High accuracy OF computation using model-based Farneback algorithm for optical camera (b) Fusing complementary advantages to mitigate accuracy-latency trade-off }

\label{fig:4}
\vspace{-0.7cm}
\end{figure}

Dynamic vision sensors (DVS) or event cameras provide a new mode of visual information with the visual data appearing as a continuous asynchronous stream of binary events instead of discrete intensity frames. An event corresponds to a change in intensity of the pixel and thus the event stream generally corresponds to moving objects in a constant light environment. Event cameras offer low power, very high dynamic range in addition to fine temporal resolution \cite{event_camera_survey}. The asynchronous event generation circumvents the fundamental speed limitation put by the frame rate. However, assigning the events to objects and thus matching the features for flow computation is complex because of the unavailability of intensity information.

Approaches to take advantage of the high throughput data can be briefly categorized into conventional model-based approaches and CNN or more recent spiking neural networks. CNNs \cite{evflow}, \cite{ecn} and hybrid approaches \cite{spikeflownet}, \cite{fusionflow} have CNN backend causing similar latencies as optical frames. Model-based approaches use iterative optimizations \cite{event_model1}, \cite{event_model2}, \cite{event_model3} causing similar processing modality as that of the optical camera and are speed limited. SNN approaches use Spike-time dependent plasticity \cite{TPAML2019}, delay coding \cite{TBioCAS2018},\cite{BioCAS2013} for training with smaller networks achieving high speed. But their accuracy is limited due to the lack of reliable training methods and are typically applied to simple custom made datasets. Thus, although event cameras have high potential speed get stuck in low throughput with conventional processing or lower accuracy with spiking networks.

This provides us with two modes of visual information namely frame and events with complementary advantages of accuracy and speed respectively (Fig. 1). Temporally detailed event stream promises high speed whereas optical frames offer high accuracy with spatial details. We envision fusing this multi-modal data to extract speed while maintaining accuracy by fusing the inferences from both pipelines. The high speed and lower accuracy event prediction is fused with a low-speed high-accuracy frame inference to induce robustness against noise while boosting the throughput. Computation on events is carried out using the shallow and local computation based leaky CNN filter that imitates correlation-based flow estimation similar to rabbit and insects \cite{rabbit_retina},\cite{fly3}. The system is validated on the MVSEC dataset \cite{mvsec} to show $19\%$ increase in error while boosting the throughput by $4\times$. The application to a rapidly moving drone flight shows that the movements that are fast to be captured by the optical camera are successfully captured by the event camera and OF is fused with high accuracy. This work demonstrates the potential of multi-modal fusion systems for overcoming trade-offs in frame-based processing.

\section{Methodology}

\subsection{Flow Estimating Leaky CNN Filter}

The OF estimation network needs to take the timestamped event stream to predict the flow at each active pixel. The shallow 3 layered leaky CNN is shown in Fig 2(a). Layer 1 acts as a leaky accumulator as shown in the equation. The most recent event adds to the current activation while the leakiness causes the contribution of previous events to diminish over time. This allows the neurons in a neighbourhood to roughly predict the direction of motion of the object in the field of view from smaller activation to higher. The accumulated activations are shown in Fig 2(b). 

The second layer computes the difference in the consecutive activation of neurons in both vertical and horizontal directions as this encodes the direction of local flow. The magnitude and polarity of the difference provide the noisy flow at the pixel. This is carried out using differential excitatory and inhibitory synapses connecting from neighbouring neurons of layer 1. Both vertical and horizontal flow are calculated separately using different kernels providing layer2a and Layer2b. Due to the high noise induced by the event camera, layer 2 activations can be seen to possess heavy granularity (Fig. 2(b)). The noise is reduced by the averaging kernel applied to layer 3 where a uniform set of excitatory synapses average the activations to provide a smoothened flow (Layer3a,3b). Each active pixel is assigned a vertical and horizontal flow value which is shown in the flow visualization.

\begin{figure}[h!]
\centerline{\includegraphics[width=0.5\textwidth,height=0.55\textwidth]{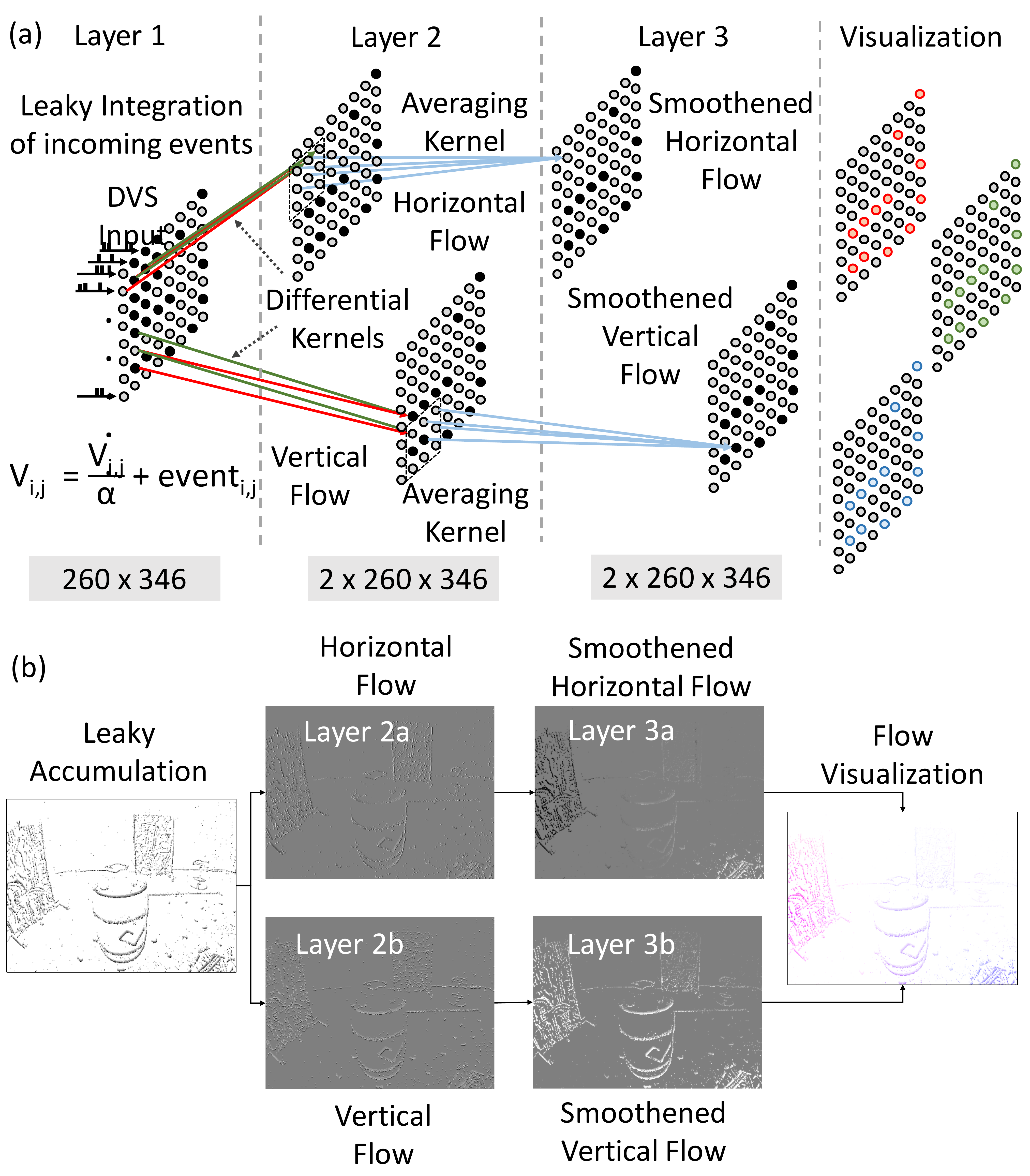}}
\vspace{-0.2cm}
\centering
\caption{(a) Leaky CNN Filter - layer 2 computes local flow and layer 3 smoothens it (b) Activations of each neuron in the layer in determining the flow}
\label{fig:4}
\vspace{-0.5cm}
\end{figure}

\subsection{Conventional Optical Flow}
The noisy flow estimation from the leaky CNN filter is to be corrected using slower but reliable conventional optical flow detectors. Multiple CNN models \cite{liteflownet, flownet} and optimization models like \cite{Farneback_flow, hs_flow} can be used. However, our dataset under consideration (MVSEC) provides grayscale optical images and DAVIS 346 experiments also provide saturated coloured images incorrectly suited for CNN models. We thus go with the conventional gradient matching Farneback algorithm \cite{Farneback_flow} because of its balance between latency and accuracy \cite{comparison_OF}. 

\subsection{Fused Event-Frame system}

\begin{figure}[h!]
\centerline{\includegraphics[width=0.45\textwidth,height=0.54\textwidth]{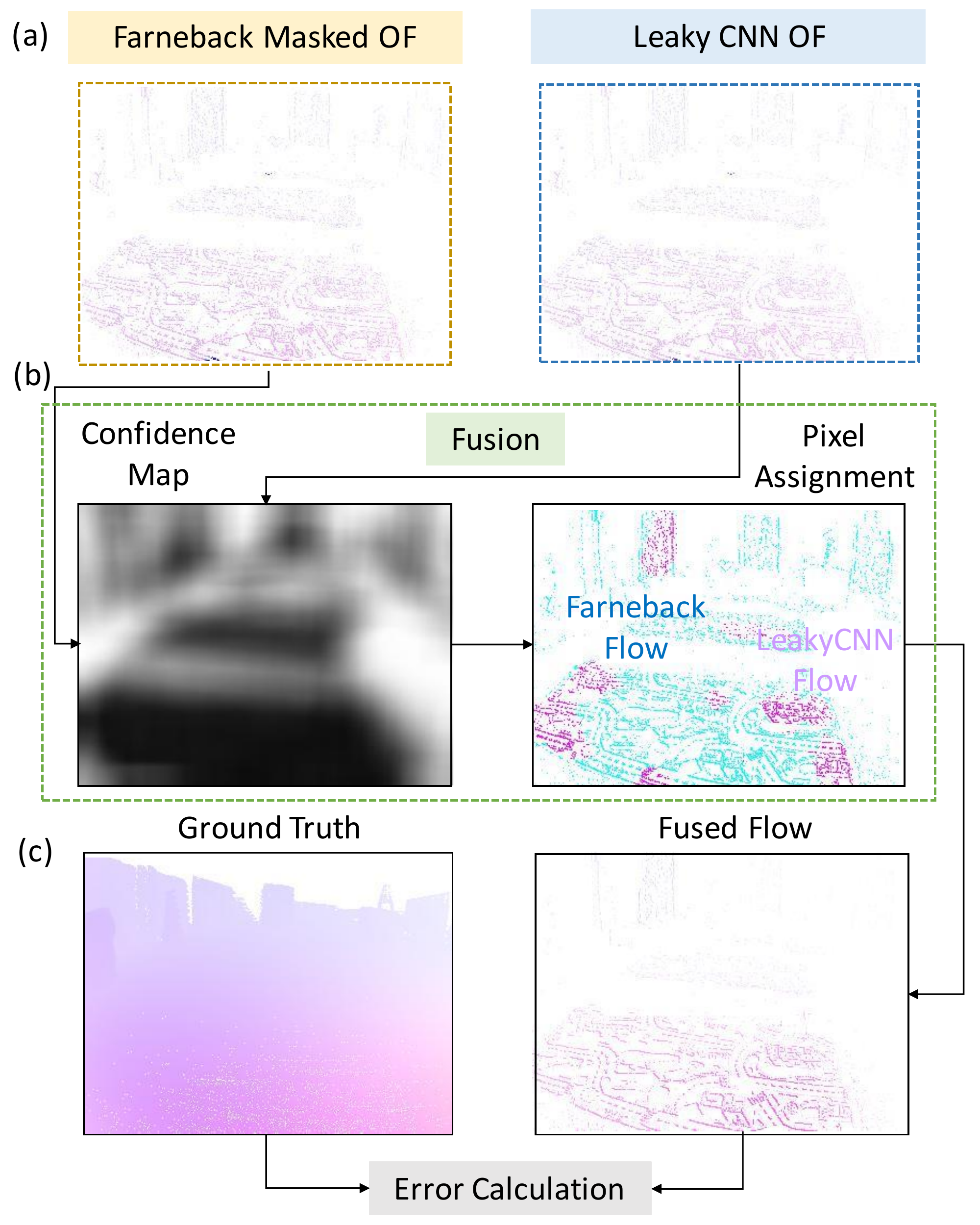}}
\centering
\caption{Flow of fusion algorithm (a) Computed flow maps by leaky filter and Farneback algorithm (b) Confidence map for event-based OF. Pixels with high confidence values are taken from here (c) Fused flow is compared with ground truth for AEE calculation}
\label{fig:4}
\vspace{-0.5cm}
\end{figure}

The fusion between the outputs from two modalities is intended to preserve the accuracy from the Farneback flow while incorporating pixels from leaky CNN that have a high likelihood of correct flow value for faster moving objects. This means that if some object has moved rapidly within the scene, the event pipeline should be able to capture that while the optical camera provides reliable detection for the background scene. This is implemented by Algorithm 1. A confidence map stores the likelihood that prediction from leaky CNN is acceptable. The pixels with high confidence scores are taken from leaky CNN prediction while others are used from previous Farneback flow computation as shown in Fig. 3.

A high confidence score is required for pixels that have seen rapid movement missed by the frame pipeline but captured by the event pipeline. The first condition requires that the leaky CNN flow prediction for the pixel differs from the flow for the same pixel in the previous frame inference (Algorithm 1 - condition 1). The second condition requires the flow to be consistent with the previous leaky CNN estimate to ensure the deviation is not because of noise and corresponds to some moving object (Algorithm 1 - condition 2). Thus the confidence score rises for a pixel when its Euclidean distance from the previous frame inference is large but close to the previous leaky CNN flow. The final fused flow map is generated by using pixels from leaky CNN inference wherever the confidence map is higher than a predefined threshold. This is outlined in Algorithm 1. Fig. 3 shows the OF values from both pipelines and the corresponding confidence map. The final fused flow is compared with ground truth to calculate the average endpoint error (AEE).

\begin{algorithm}[]
\caption{Fusion Algorithm}

\While{True}{    
    Condition 1: \\
    $Distance_{Farneback}$ = $|| OF_{Farneback}$, $OF_{leakyCNN} ||$ \\
    \eIf {$Distance_{Farneback}$ $>$ $thresh_{Farneback}$} {
    	$Error_{Farneback}$ = 1
    }{
    	$Error_{Farneback}$ = 0
    }
    
    Condition 2: \\
    $Distance_{leakyCNN}$ = $|| OF_{leakyCNN_t}$, $OF_{leakyCNN_{t-1}} ||$ \\
    \eIf {$Distance_{leakyCNN}$ $<$ $thresh_{leakyCNN}$}{
    	$Error_{leakyCNN}$ = 1
    }{
    	$Error_{leakyCNN}$ = 0
    }

    $belief$ = $Error_{Farneback}$ * $Error_{leakyCNN}$\\
    $belief$ = $belief_{prev}$ + $belief$\\
    $Confidence\ Map = Confidence\ Map + belief$ \\ 
    \eIf {Confidence Map $>$ thresh}{
    	$Pixels_{leakyCNN}$ = 1
    }{
    	$Pixels_{Farneback}$ = 1 
    }
    $OF_{fused} = OF_{Farneback} * Pixels_{Farneback}$ \\ $+ OF_{leakyCNN} * pixels_{leakyCNN}$
    
}
* is element-wise multiplication

\end{algorithm}
\vspace{-0.2cm}

\section{Results}

\subsection{Fusion Mechanism}
We begin by exploring the parametric dependence of user-defined parameters on the fusion algorithm. The fraction of pixels coming from event pipeline prediction (shown as event percent in Fig. 4) OF depends crucially upon the error tolerance thresholds (condition 1,2). Low $thresh_{Farneback}$ and high $thresh_{leakyCNN}$ results in a larger fraction of pixels coming from the event pipeline. This is depicted in Fig. 4 (a). The percentage of event OF monotonically rises when these two conditions are met and provides the user with control knobs to tune the algorithm. 

\begin{figure}[h!]
\centerline{\includegraphics[width=0.5\textwidth,height=0.2\textwidth]{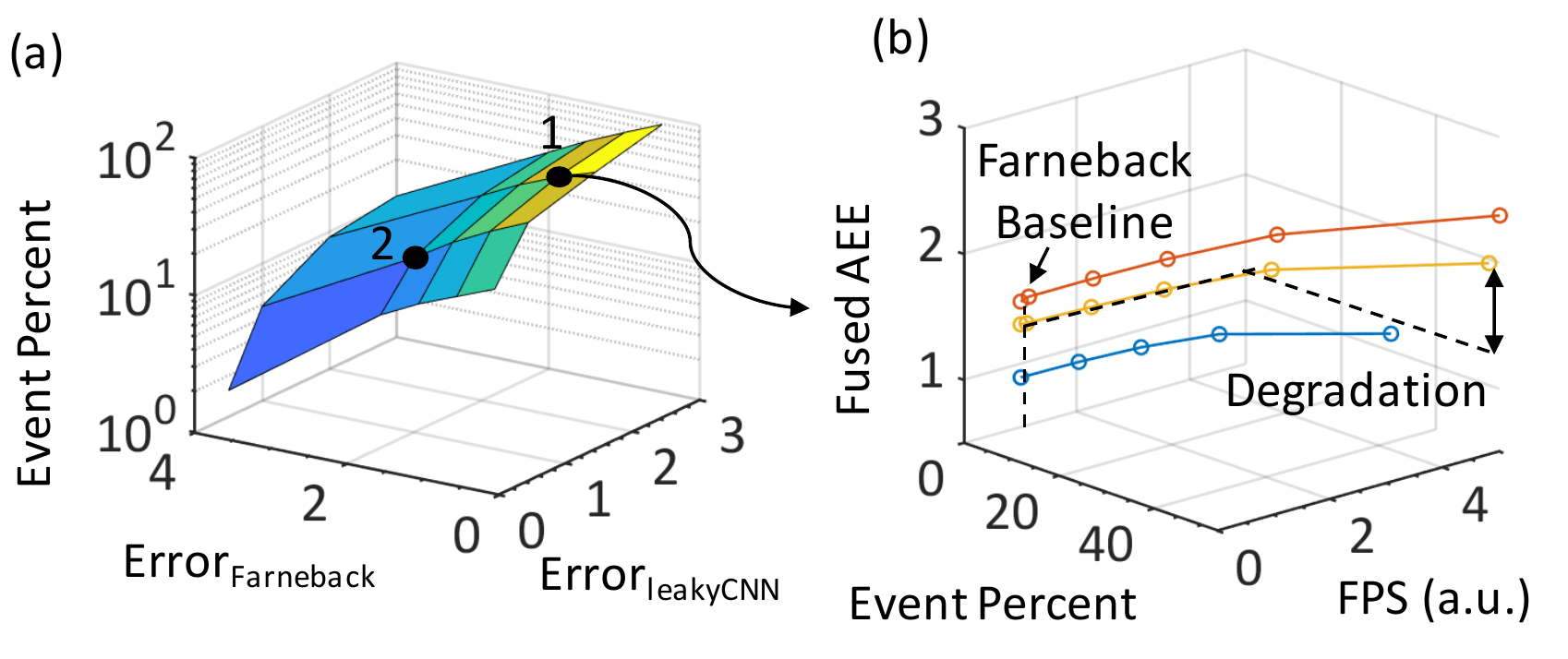}}
\vspace{-0.3cm}
\centering
\caption{Fusion parameters sweep. (a) Low $thresh_{Farneback}$ and high $thresh_{leakyCNN}$ allows high percentage of pixels from event camera pipeline in the flow (b) Accuracy gradually degrades with higher percentage of pixels from leaky CNN at a higher FPS rate}
\label{fig:4}
\vspace{-0.5cm}
\end{figure}

\begin{table}
\caption {MVSEC - Average End-Point Error (AEE) vs FPS} 
\vspace{-3mm}
\begin{center}
\begin{tabular}{| c | c | c | c | c | c |   }
\hline

\textbf{Network} & \textbf{Indoor} & \textbf{Indoor} & \textbf{Indoor} & \textbf{Time} & \textbf{FPS} \\
\textbf{}  & \textbf{Flying 1} & \textbf{Flying 2} & \textbf{Flying 3} & \textbf{ms} & \textbf{}  \\

\hline
\hline

Unflow \cite{unflow} & 0.5 & 0.7 & 0.55 & - & - \\
\hline
EV Flow \cite{evflow} & 1.03 & 1.72 & 1.53 & 48 & 21  \\
\hline
SpikeFlow \cite{spikeflownet} & 0.84 & 1.28 & 1.11 & 23.11 & 43 \\
\hline
Zhu et. al. \cite{cvpr2019} & 0.58 & 1.02 & 0.87 & - & - \\
\hline
FusionFlow \cite{fusionflow} &  0.56 & 0.95 & 0.76 & - & - \\
\hline
Full ANN \cite{fusionflow} &  0.68 & 0.97 & 0.97 & - & - \\
\hline
ECN \cite{ecn} &  0.49 & 0.43 & 0.48 & 4 & 250 \\
\hline
This work &  0.95 & 1.55 & 1.38 & 24  & $>$ 41  \\
\hline

\end{tabular}
\end{center}
\vspace{-0.5cm}
\end{table}

A high percentage of pixels coming from the event pipeline is expected to corrupt the accuracy because of the noise it injects. As the FPS increases, more frames from the event pipeline are processed between every inference of the frame pipeline. This causes the percentage contributed from event pipeline prediction to increase as shown in Fig 4(b). The error thresholds are from point `1' in Fig. 4(a). The error monotonically rises with FPS and percentage contributed from the event pipeline.

A comparison of this method with previous methods for the MVSEC dataset that captures multiple indoor scenes with an event camera mounted on a drone is carried out in Table 1. Error thresholds are from point `2' in Fig 4(a). The FPS of the Farneback flow is taken as the baseline of 1 while the event stream from the event camera between the consecutive frames is divided into multiple frames and is processed through the event pipeline. Fig. 5(a) shows that as the FPS rises, the noisy contribution of the event pipeline rises. We observe that the error degradation from Farneback to the fused method is small while the frame rate increases significantly (4x). Thus, the fusion algorithm mitigates the trade-off between event and frame pipelines extracting their complementary advantages. The throughput vs. FPS trade-off of various previous methods is shown in Fig. 5(b). The FPS shown for our method is for pipelined execution in FPGA as described in the next section.

\vspace{-3mm}
\begin{figure}[h!]
\centerline{\includegraphics[width=0.5\textwidth,height=0.23\textwidth]{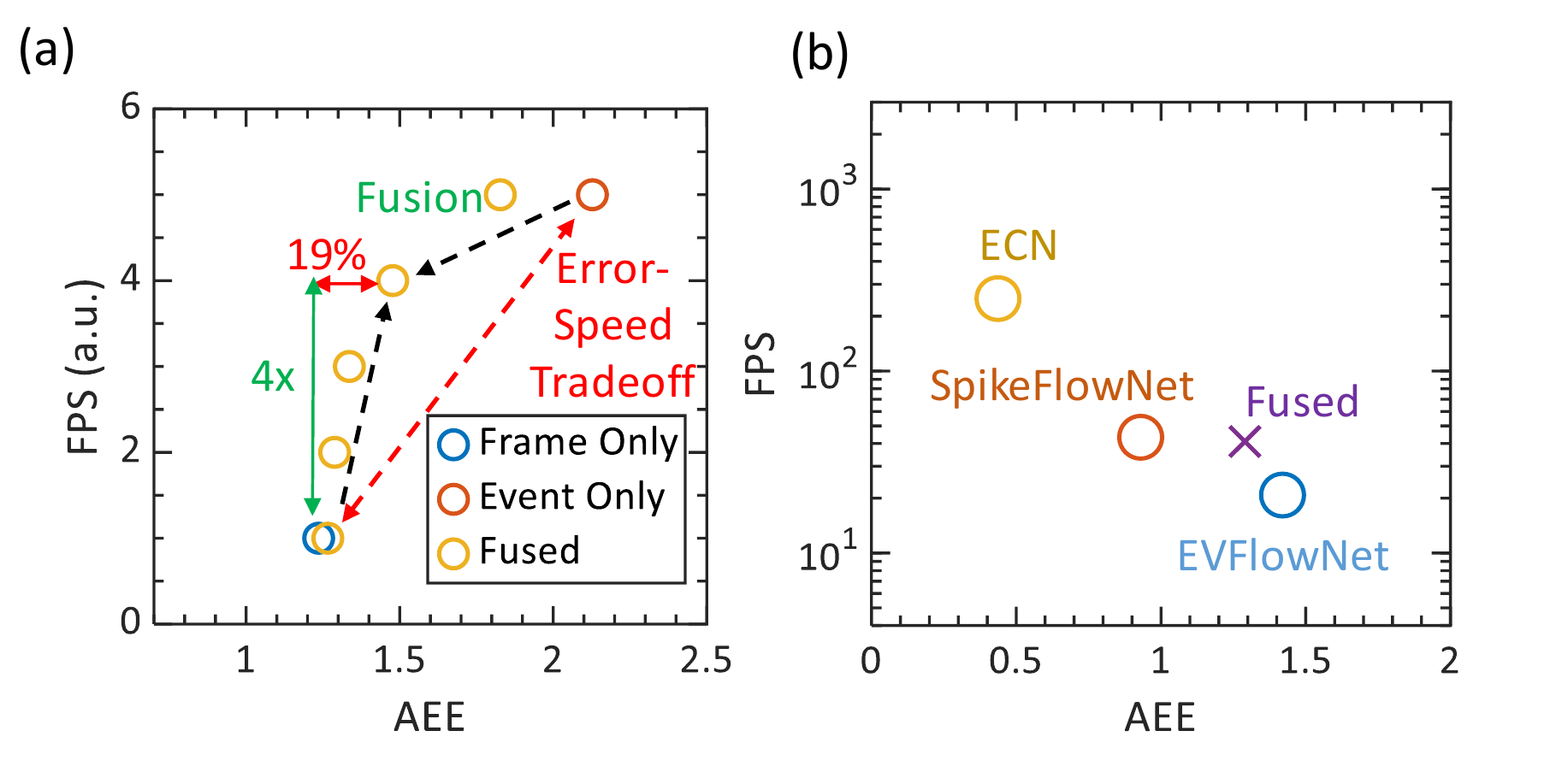}}
\centering
\vspace{-0.5cm}
\caption{(a) Overcoming the accuracy latency trade-off (b) Comparison with previous approaches }
\label{fig:4}
\vspace{-0.5cm}
\end{figure}

\subsection{Real World Demonstration}
\vspace{-0.4cm}
\begin{figure}[h!]
\centerline{\includegraphics[width=0.45\textwidth,height=0.36\textwidth]{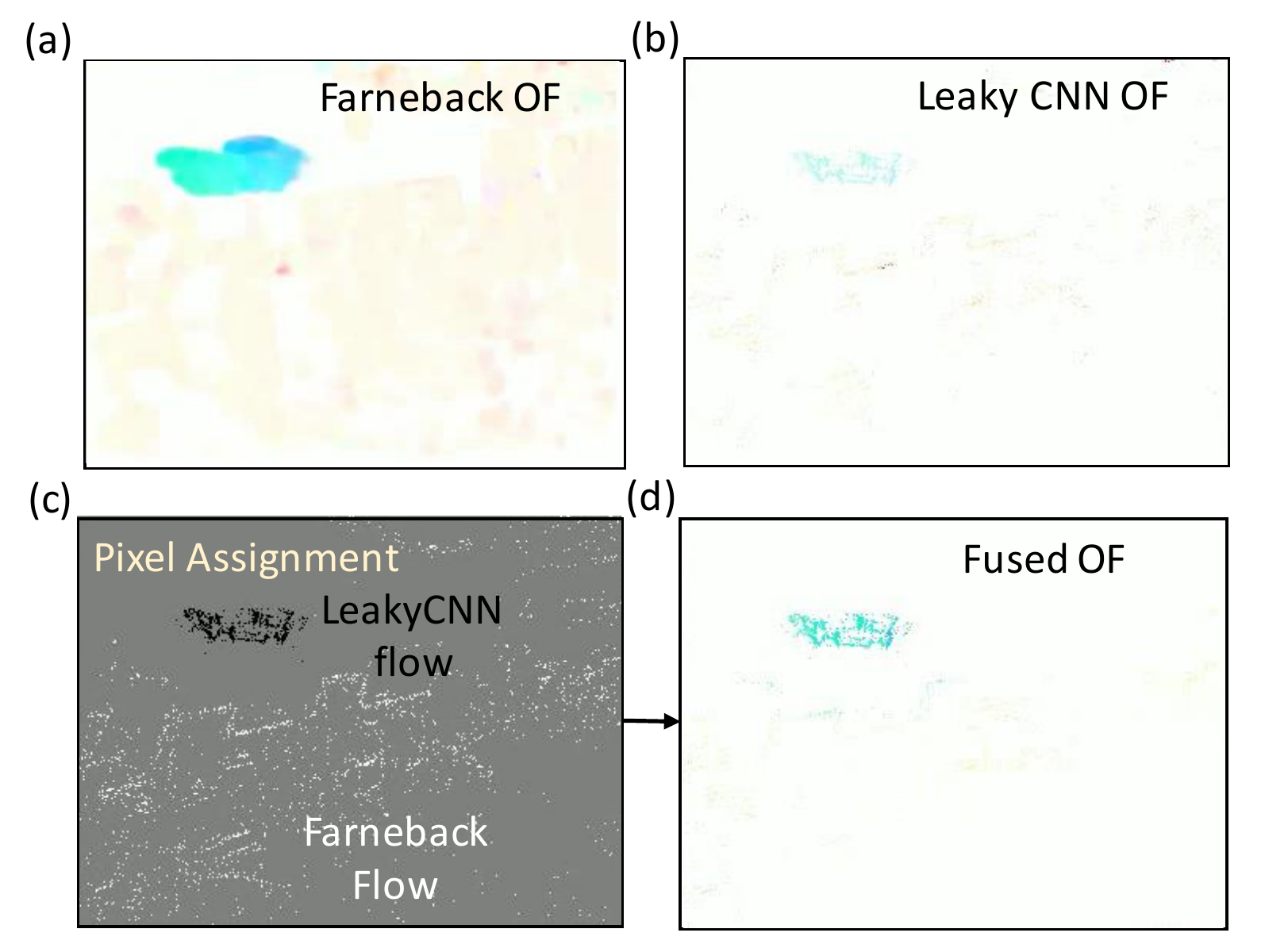}}
\centering
\caption{Screenshots from drone experiment with high-speed flow for drone fused from leakyCNN while the background is taken from Farneback flow}
\vspace{-3mm}
\label{fig:4}
\end{figure}

The system is applied to a real-world scenario of a flying drone being captured with a handheld DAVIS346. This provides both color and frame information. The fused optical flow method is applied to this where the rapid movement of the drone that is harder to capture using the only optical camera can be correctly identified using the event-camera pipeline. A screenshot from the processing is shown in Fig. 6. The drone can be seen to be having a high confidence score while the background information is inserted from the Farneback inference. The link to the video is available at (\href{https://www.youtube.com/watch?v=O587-hzIDwM}{\underline{\color{blue}demo-1}})\footnote{https://www.youtube.com/watch?v=O587-hzIDwM}.

\subsection{Throughput Estimation}

We estimate the hardware consumption of the algorithm in a pipelined synchronous execution on FPGA using vitis high level synthesis 2021. Leaky CNN and fusion algorithm is executed on the FPGA while Farneback inference is assumed to be fed externally by a conventional processor. Assuming 10000 events are processed in every run, $\sim$ 3 million operations are required per prediction. The synthesis is for Xilinx Virtex UltraScale family FPGA chip xcvu125-CIV-flvd1517-3-e. $8.09$ $\times$ $10^6$ clock cycles are consumed in generating and fusing one leakyCNN prediction which for a 333 MHz clock promises 41 FPS. 10 DSP(0.83\%), 5621 FF(0.39\%), 7957 LUT(1.11\%) and 1106 BRAM(43.88\%) is consumed in the execution highlighting the potential for resource-constrained edge-application. The throughput is currently limited by the number of memory ports and with the incorporation of tiling and event-based hardware design, the latency can be improved further.
\vspace{-0.1cm}
\section{Discussion}

Our leaky CNN takes inspiration from the correlation type flow estimation model proposed for animal brains \cite{bio_review1}. Spatial delay of the motion response is encoded by the leaky accumulator while the differential synaptic kernel in layer 2 adds direction sensitivity. Many recent works observe \cite{fly1, fly2} and also individually map \cite{fly3} the direction sensitive activation of visual neurons for flies. Similar behaviour was observed in rabbits \cite{rabbit_retina}. The network may be made more noise-tolerant with additional kernels sensitive to intermediate angles and multi-synaptic kernels instead of the differential kernels presented here.

Previous works have also explored multi-modal systems of optical flow computation. Fusion-flowNet \cite{fusionflow} fuses SNN and CNN activations while training them as a single network. \cite{fused_pedestrian_detection} used both frame and event data as input to a CNN and train them simultaneously. The key difference lies in the fact that we use a bio-inspired neuronal filter to build a processing pipeline for the event camera and the fusion happens in the final stage. This for the first time to the best of our knowledge uses independent pipelines to take their complementary advantages independently without any composite training.

\vspace{-0.1cm}
\section{Conclusion}
We proposed a fusion system for frame and event cameras for high-speed optical flow detection. Our network imitates some characteristics of biological neuronal processing and combines complementary speed and accuracy advantages of the two vision systems. The system is validated on the MVSEC dataset and subsequently applied to high-speed drone motion to demonstrate a real-world application. This shows that the accuracy latency trade-off of frame-based processing can be mitigated by using input and processing frameworks from other modalities.

\vspace{-0.1cm}
\section{Acknowledgement}
\vspace{-0.2cm}
This work was supported by IARPA sponsered Microelectronics for AI program

\bibliographystyle{ieeetr}
\bibliography{ref}

\end{document}